\documentclass[sigconf]{acmart}

\usepackage{booktabs} 

\setcopyright{rightsretained}

\acmDOI{}

\acmISBN{}

\acmConference[]{}
\acmYear{}
\copyrightyear{}

\acmPrice{}

\begin{document}
\title{Population network structure impacts genetic algorithm optimisation performance}

\author{Aymeric Vi\'{e}}
\orcid{0000-0002-7178-1380}
\affiliation{%
  \institution{Mathematical Institute, University of Oxford}
  \institution{Institute of New Economic Thinking, University of Oxford}
  \city{Oxford} 
  \country{United Kingdom}
}
\email{vie@maths.ox.ac.uk}
\renewcommand{\shortauthors}{Aymeric Vi\'{e}}

\begin{abstract}
A genetic algorithm (GA) is a search method that optimises a population of solutions by simulating natural evolution. Good solutions reproduce together to create better candidates. The standard GA assumes that any two solutions can mate. However, in nature and social contexts, social networks can condition the likelihood that two individuals mate. This impact of population network structure over GAs performance is unknown. Here we introduce the Networked Genetic Algorithm (NGA) to evaluate how various random and scale-free population networks influence the optimisation performance of GAs on benchmark functions. We show evidence of significant variations in performance of the NGA as the network varies. In addition, we find that the best-performing population networks, characterised by intermediate density and low average shortest path length, significantly outperform the standard complete network GA. These results may constitute a starting point for network tuning and network control: seeing the network structure of the population as a parameter that can be tuned to improve the performance of evolutionary algorithms, and offer more realistic modelling of social learning.
\footnote{All source code for the NGA, the figures and the results are available at \url{https://github.com/aymericvie/networked-genetic-algorithms}.}
\end{abstract}



\maketitle

\section{Introduction}

A genetic algorithm (GA) is a search method that optimises a population of individuals by simulating natural evolution \cite{holland1992genetic,vie2020qualities} and reproduction of the fittest. Individuals in the population are assumed to be able to mate with any other. Such a connection between all individuals can be seen in a network science perspective as a constant, fully connected network in the population. However, in social learning as in nature, populations are not fully connected, as individuals only interact with a finite subset of the whole \cite{vasques2020transitivity}. The network structure of these interactions have different structures, that can be an important determinant of population-level dynamics \cite{vie2019information}, and accounts in nature for changes in population genetic diversity and the emergence of different species \cite{broquet2010genetic}. While network design in operations research \cite{owais2018complete} or neural architecture search \cite{lu2018nsga} are active areas of research, there have been no attempts so far to investigate alternative GA network architectures. \\

It is unknown whether and to what extent this assumption of completeness and the population network structure impact GAs performance. In this article, we introduce \textit{Networked GAs} (NGA) that use alternative population network structures, and measure their performance over benchmark optimisation tasks to identify the impact of network architectures. We consider various population structures generated by Erdos-R\'{e}nyi random networks \cite{erdos1960evolution} and Albert-Barabasi scale-free networks \cite{barabasi1999emergence}.\\

We show evidence of significant variations in performance of the NGA as the network structure varies. We compare optimisation performance to various network metrics such as \textit{density}, which measures the ratio of realised links to the number of possible links; and \textit{average shortest path length}, which describes the average minimum number of links to be crossed to connect any pair of individuals. NGA performance is higher with intermediate levels of density and low average shortest path length, settings that favor circulation of fittest genomes while preventing those fittest solutions from becoming dominant too fast. In addition, we discover that the optimal network NGA outperforms the standard GA with a fixed complete network structure. \\

The results open the possibility of \textit{network tuning}: identifying the optimal population network architecture to maximise the GA optimisation performance. It is also possible that different network structure could be optimal during the run. \textit{Network control}, i.e. allowing this population network to change over time, could further improve the performance of GAs, model social learning and make one step towards a closer implementation of biological features in evolutionary algorithms \cite{miikkulainen2021biological}.

\begin{figure*}[ht!]
\includegraphics[width=0.8\textwidth]{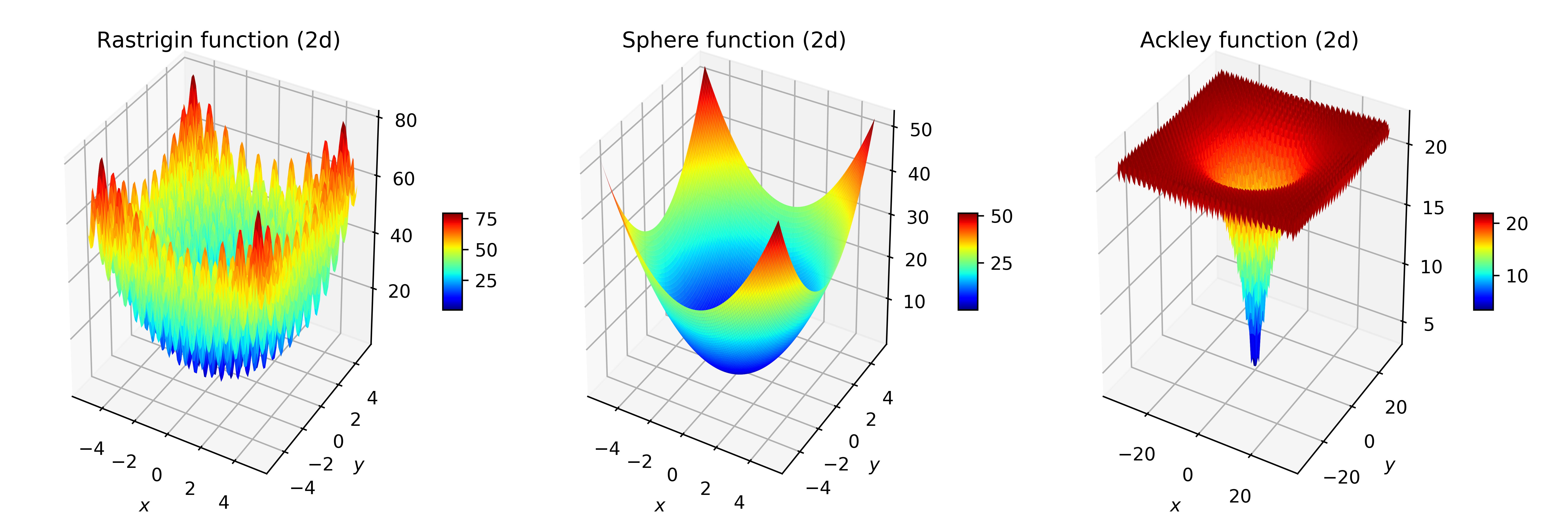}
\caption{Fitness landscapes of the three test functions (Rastrigin, Sphere, Ackley) with $d=2$.}
\end{figure*}

\section{Algorithm}

\subsection{Test functions}
To compare the performance of the GA with respect to the population network structure, we choose three popular test functions as benchmarks: the Rastrigin function (\ref{rastrigin}), the Sphere function (\ref{sphere}), and the Ackley function (\ref{ackley}), with respective domains in Table \ref{function_domain} and dimensionality $d=2$. Their global optimum is at $f(\textbf{0})=0$.

\begin{equation}
\label{rastrigin}
    f_1(\textbf{x}) = 10d + \sum_{i=1}^d \left(x_i^2 - 10\cos{2\pi x_i}\right)
\end{equation}

\begin{equation}
\label{sphere}
    f_2(\textbf{x}) = \sum_{i=1}^d x_i^2
\end{equation}

\begin{equation}
\label{ackley}
    f_3(\textbf{x}) = -20\exp{\left(-0.2\sqrt{\frac{1}{d}\sum_{i=1}^dx_i^2}\right)} - \exp{\left(\frac{1}{d}\sum_{i=1}^d\cos{2\pi x_i}\right)} + e + 20
\end{equation}

\begin{table}
  \caption{Domains of test functions}
  \label{function_domain}
  \begin{tabular}{ccc}
    \toprule
    Function & Name & Domain\\
    \midrule
    $f_1$ & Rastrigin function & $-5.12 \leq x_{1\leq i \leq d} \leq 5.12$\\
    $f_2$ & Sphere function & $-5.12 \leq x_{1\leq i \leq d} \leq 5.12$\\
    $f_3$ & Ackley function & $-32.768 \leq x_{1\leq i \leq d} \leq 32.768$\\
  \bottomrule
\end{tabular}
\end{table}

\subsection{The networked-population genetic algorithm}

\subsubsection{Genetic representation and sampling}

To optimise the test functions with dimensionality $d$, each individual is a string of $d$ real numbers sampled  uniformly in the corresponding search domain of Table \ref{function_domain}. A total of $n$ individuals are generated.

\subsubsection{Genetic operators}

For each individual $i$, the resulting value $f(i)$ of the test function is computed. As the global minima of our test functions are equal to 0, we strive to minimise $f(i)^2$. To create two offspring, the first individual $i$ is chosen with a fitness-proportionate method, with selection probability:

\begin{equation}
    \label{selection_probability_full_population}
    \pi_i = \frac{\hat{f}(i)}{\sum_{j=1}^n\hat{f}(i)^2} \ \ \ \text{ with } \hat{f}(i)=\frac{1}{1+f(i)}
\end{equation}

The networked-population GA (NGA) differs from the standard GA in the selection of individuals to mate. If individual $k$ has been selected in the population, the second will be selected among the $N(k)$ individuals sharing a link with individual $k$, instead of performing again selection on the full population as in the standard GA. Probabilities of selection are identical to equation \ref{selection_probability_full_population} applied to the subset $N(k)$ rather than the full population size $n$.


Once the two parents have been selected, uniform crossover is performed with probability $\rho$. A crossover point in $[0,d]$ will be randomly and uniformly determined, and chromosomes of parents will be exchanged after the crossover point to generate two offspring. With probability $1 - \rho$, the individuals instead create exact copies of themselves. Finally, random Gaussian mutations occur with probability $\mu$, and change the value of the mutated element by $\epsilon \sim \mathcal{N}(0,1)$. The algorithm runs for $\tau$ time periods. Standard parameter values are used for the NGA, and presented in Table \ref{parameter-configuration}.



\subsubsection{The population network structure}

In the canonical genetic algorithm, the crossover operator can involve any combination of individuals in the population, corresponding to complete network NGA. In the NGA, we consider Erdos-R\'{e}nyi (ER) random networks and Albert-Barabasi (AB) scale-free networks. ER graphs are described by the link formation probability $p$, from $0$ (empty network, or island GA model without migration) to $1$ (complete network, standard GA). 
AB graphs are described by the intensity of the preferential attachment $m$, generating tree networks for low values of $m$, then graphs with multiple hubs, up to star networks for highest values of $m$. Figure 2 shows examples of the generated networks. As these two network parameters change, so do the corresponding network features. ER networks become connected (i.e. there exists a path between any pair of nodes) at $p=\frac{\ln n}{n}$ \cite{erdos1960evolution}, and increase monotonically in density with $p$. In AB graphs, $m$ determines the level of density, and the average shortest distance between any pair of nodes \cite{barabasi1999emergence}\footnote{We refer the reader to the GitHub repository for a graphical representation of the impact of network parameters $p$ and $m$ over network features.}. This population network structure is drawn once at the start of the run, and left constant for its duration. 

\begin{table}
  \caption{GA parameter configuration}
  \label{parameter-configuration}
  \begin{tabular}{ccl}
    \toprule
    Parameter & Meaning & Value\\
    \midrule
    $n$ & Population size & 50 \\
    $\rho$ & Crossover rate & 0.7 \\
    $\mu$ & Mutation rate & 0.05 \\
    $\tau$ & Number of iterations & 100 \\
  \bottomrule
\end{tabular}
\end{table}

\begin{figure}[ht!]
    \centering
    \includegraphics[width=0.45\textwidth]{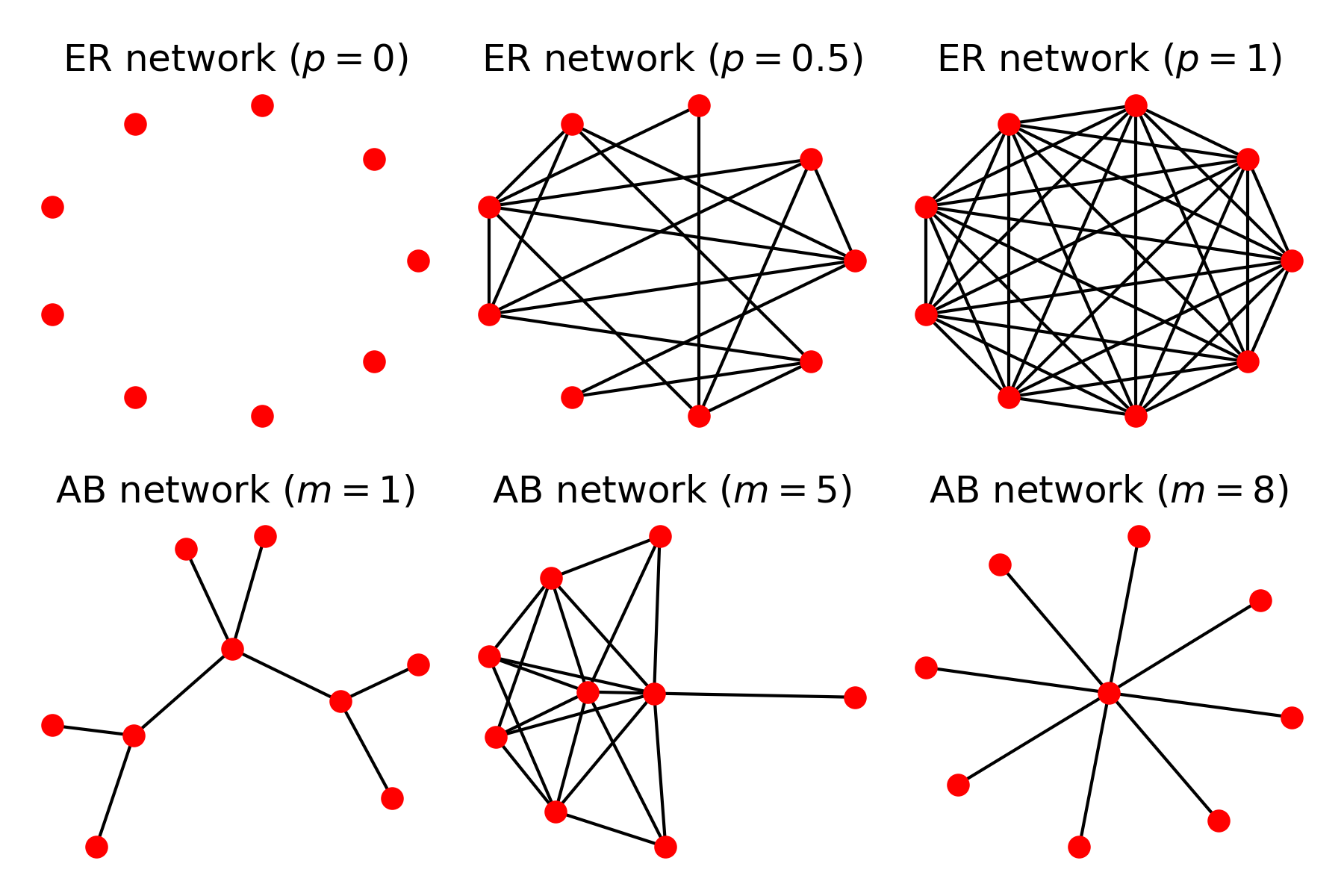}
    \caption{6 examples of Erdos-R\'{e}nyi (ER, \cite{erdos1960evolution}) and Albert-Barabasi (AR, \cite{barabasi1999emergence}) networks with 10 nodes}
\end{figure}

\section{Results}

\subsection{NGA performance over iterations}

\begin{figure}[ht!]
    \centering
    \includegraphics[width=0.5\textwidth]{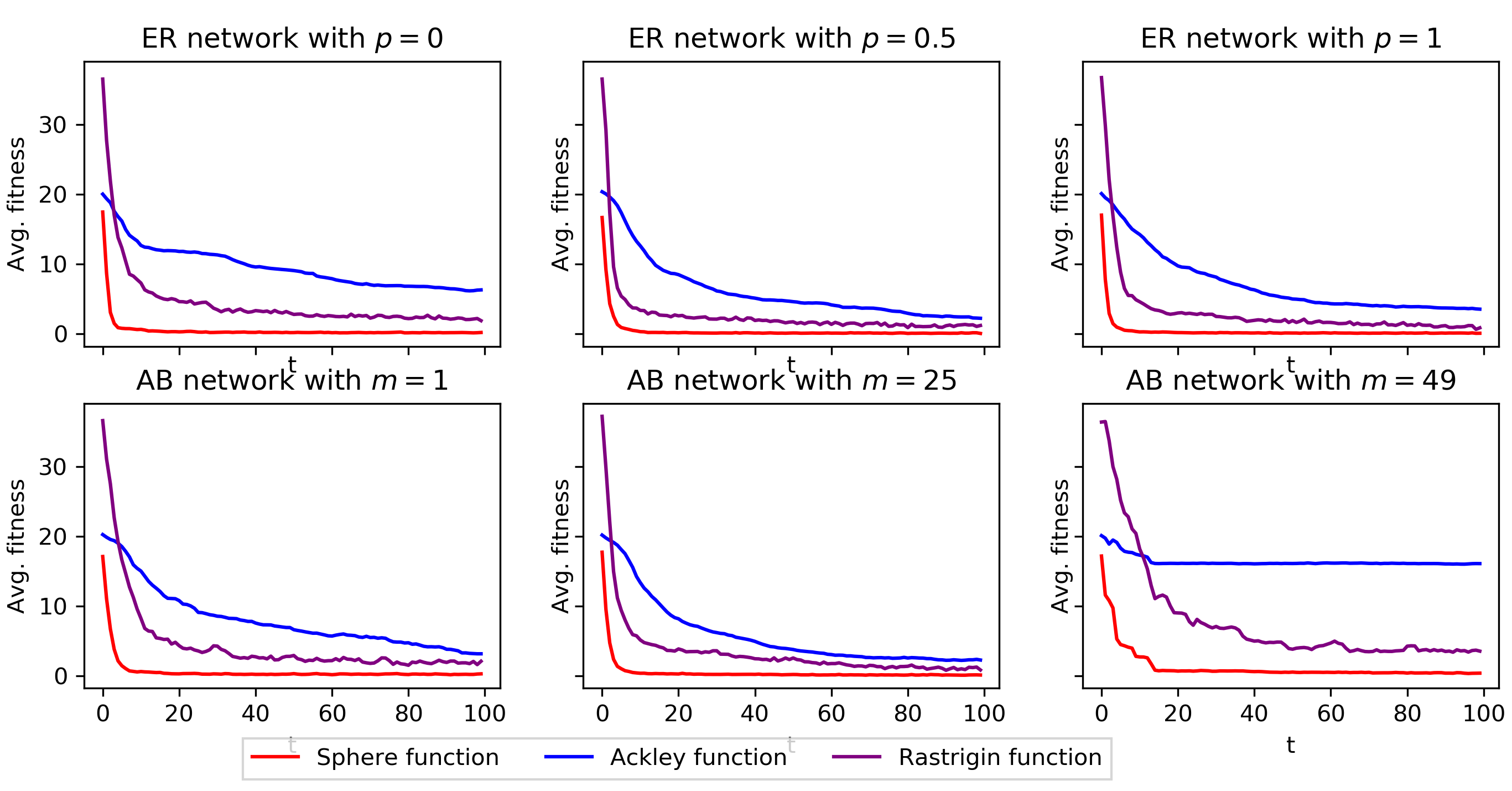}
    \caption{Average fitness for the three test functions as a function of time, for various network types and parameters, with 10 repetitions}
\end{figure}

Various network structures visibly affect the performance of the GA in Figure 3. Average fitness at the final iteration $\tau = 100$ and convergence speed vary with the network type (random or scale-free) and parameters ($p$ and $m$). In particular, the empty network (top left of Figure 2) and the star network (bottom right) achieve a higher fitness score than their counterparts, showing evidence for weaker performance in all benchmark tasks, with differences attributed to the respective task difficulty. Comparing the performance of the complete network (top right) commonly used in GAs with other structures suggests that alternative network structures may offer better optimisation performance. Notably, optimising the Ackley function appears to be more successful in an ER graph with $p=0.5$ and with the AB graph with $m=25$. These particular results invite us to analyse more comprehensively how benchmark performance varies with the link formation probability $p$ in ER networks, and with the preferential attachment factor $m$ in AB graphs. 

\subsection{The impact of network structure over performance}

\begin{figure}[ht!]
    \centering
    \includegraphics[width=0.5\textwidth]{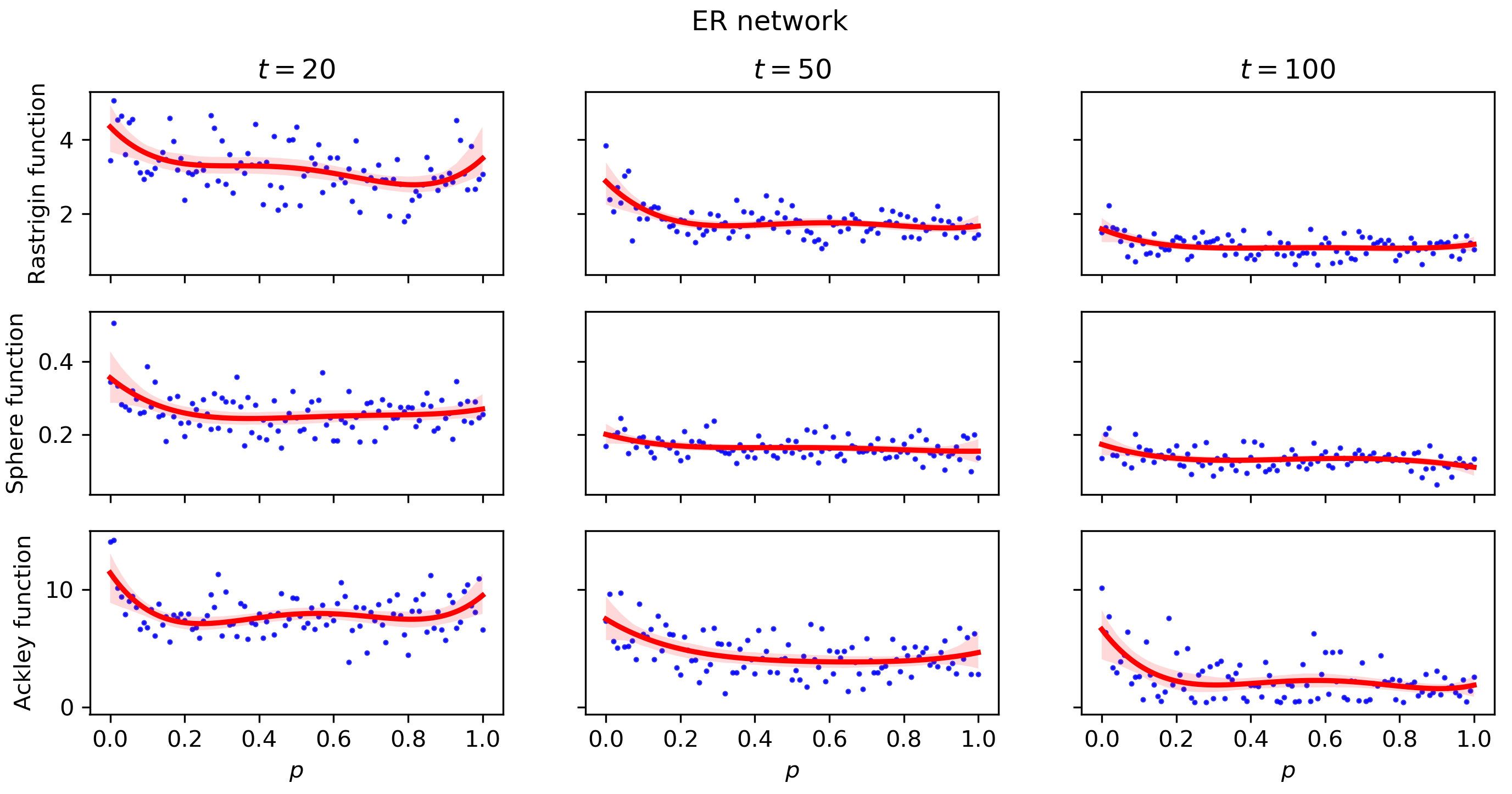}
    \caption{Average fitness for the test functions optimisation in ER networks with link probability $p$}
\end{figure}

We first evaluate how the link formation probability $p$ in Erdos-R\'{e}nyi networks changes NGA performance, shown in Figure 4. For each value of $p \in [0,1]$ with an increment of $0.01$, we run the algorithm $10$ times and record the average fitness performance at $t=20$, $t=50$ and $t=100$. As the population network becomes connected at $p=0.046$, a significant improvement of performance is observed. Below this threshold, as crossover is heavily limited by the disconnected network, most of the evolution happens through the mutation operator only, limiting the performance of the NGA. Above the connectedness threshold, the average fitness moderately decreases, exhibiting peaks of lower average fitness in the range $p\in[0.2,0.3]$ and $p\in[0.8,0.9]$. The polynomial fits (order 4) suggest that such intermediate levels of $p$ allow a better performance than the complete network. \\

We then study the influence of the preferential attachment factor $m$ in Albert-Barabasi scale-free networks. $m$ varies from $1$ to $49$ with an increment of $1$. Tree networks (low $m$), and star networks (highest $m$) perform significantly worse in all benchmark tasks, as shown by Figure 5. Tree networks are indeed connected, but exhibit high shortest path lengths, limiting the circulation of good solutions in the population. Star networks are vulnerable to errors or premature local dominance of the central hub. Polynomial fits suggest again the existence of two performance peaks for values of $m$ close to 10, and to 30, that coincide with intermediate levels of network density but low average shortest paths lengths, similar to biological networks' characteristics  \cite{huizinga2016does}. 

\begin{figure}
    \centering
    \includegraphics[width=0.5\textwidth]{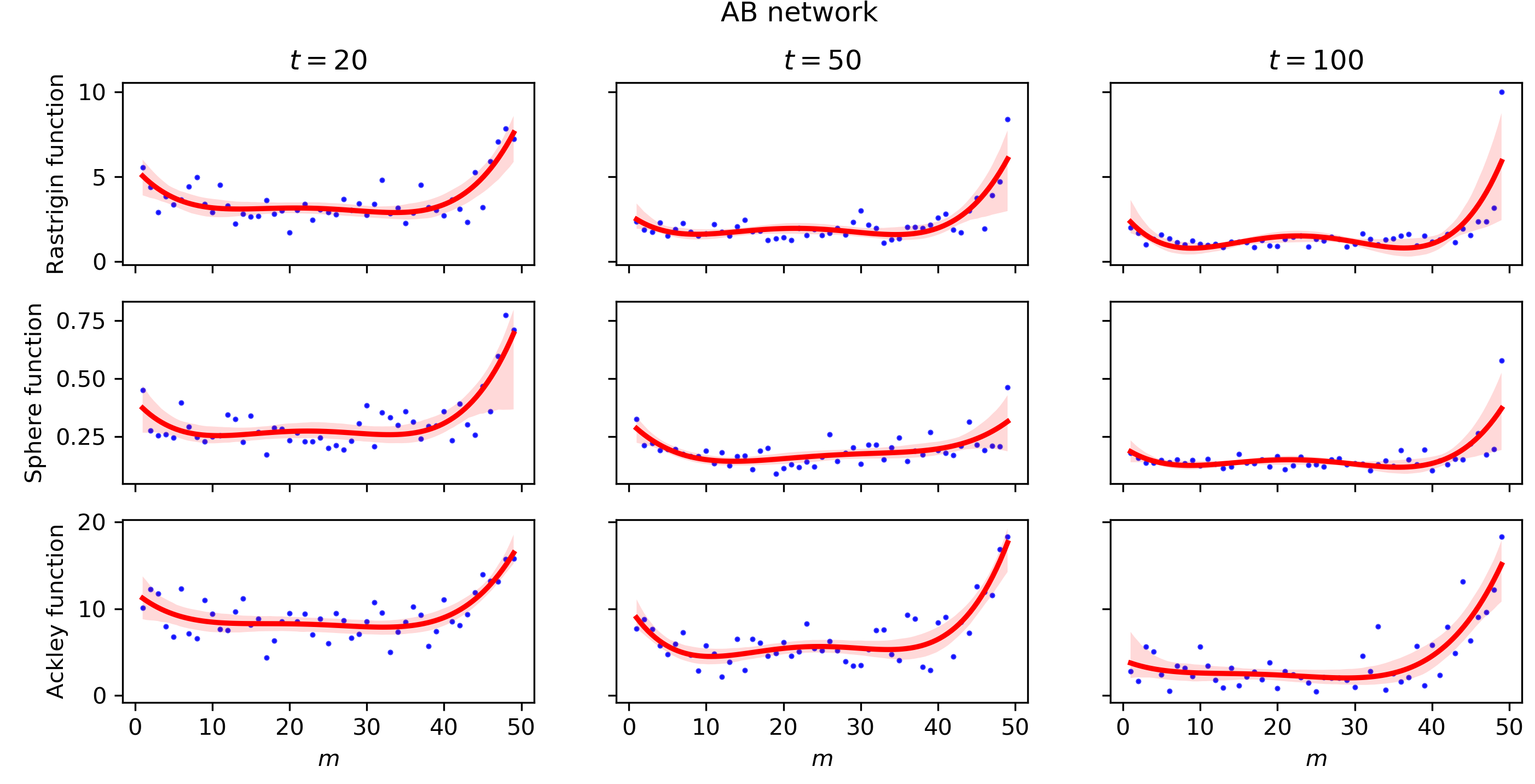}
    \caption{Average fitness for the test functions optimisation in AB networks with preferential attachment factor $m$}
\end{figure}

\begin{table*}[ht!]
  \caption{Average fitness in the standard GA (GA) and the best average fitness of the NGA across all networks (ER$^*$ \& AB$^*$)}
  \label{comparison}
  \begin{tabular}{c|ccc|ccc|ccc}
    \toprule
    Function \& Time & \multicolumn{3}{c}{$\tau=20$} & \multicolumn{3}{c}{$\tau=50$} & \multicolumn{3}{c}{$\tau=100$}\\
    \midrule
    Rastrigin & 3.067 & 1.796 & \textbf{1.705} & 1.445 & \textbf{1.072} & 1.093 & 1.043 & \textbf{0.625} & 0.82 \\
    Sphere    & 0.257 & \textbf{0.164} & 0.172 & 0.138 & 0.1 & \textbf{0.09} & 0.135 & \textbf{0.064} & 0.103 \\
    Ackley    & 6.597 & \textbf{3.859} & 4.328 & 2.809 & \textbf{1.171} & 2.157 & 2.568 & \textbf{0.412} & 0.451 \\
    \midrule
    Network   & GA & ER$^*$ & AB$^*$ & GA & ER$^*$ & AB$^*$ & GA & ER$^*$ & AB$^*$\\
  \bottomrule
\end{tabular}
\end{table*}

\subsection{Network tuning outperforms the standard GA}

The optimal-network NGA significantly outperforms the standard GA in all benchmark tasks. The best alternative NGA (ER or AB) exhibits average fitness scores that are 53\% lower than those obtained by the standard GA (Table \ref{comparison}). Intermediate ER networks are often the best performing structure. The best solutions' potential for dominance and premature convergence to local optima is limited by the network incompleteness, while the dense network structure still allows fittest individuals to circulate. This translates into particularly high performance gains in multimodal landscapes, e.g. the Ackley function.
Likewise optimisation of neural network architectures \cite{stanley2002evolving}, the GA population networks may be tuned to improve GAs performance. Though robust across our three test functions, repeated runs and various time horizons, future investigation with alternative configurations to Table \ref{parameter-configuration} and test functions could support these findings. \textit{Network tuning} (architecture optimisation before the run), or \textit{network control} (during the run) can unlock new performance in population-based search algorithms for optimisation and social learning tasks. Further research could explore what evolution mechanisms could converge to optimal population network structures.

\section{Conclusions}

We introduced Networked GAs, that constrain crossover in a network. We considered varied population network structure in random and scale-free graphs. Using NGAs to optimise the Ackley, Rastrigin and Sphere test functions, we found evidence of significant changes in the average population fitness at given iterations as the population network structure varies. The best-performing networks significantly outperform the standard complete network GA in all benchmark tasks, suggesting that the population network structure, like other GA parameters, could be tuned to improve performance. 

\begin{acks}
The author thanks Doyne J. Farmer, Alissa M. Kleinnijenhuis, Renaud Lambiotte, José Moran and Manpreet Singh for precious support and advice.

\end{acks}

\bibliographystyle{ACM-Reference-Format}
\bibliography{main} 


\begin{thebibliography}{12}


\ifx \showCODEN    \undefined \def \showCODEN     #1{\unskip}     \fi
\ifx \showDOI      \undefined \def \showDOI       #1{#1}\fi
\ifx \showISBNx    \undefined \def \showISBNx     #1{\unskip}     \fi
\ifx \showISBNxiii \undefined \def \showISBNxiii  #1{\unskip}     \fi
\ifx \showISSN     \undefined \def \showISSN      #1{\unskip}     \fi
\ifx \showLCCN     \undefined \def \showLCCN      #1{\unskip}     \fi
\ifx \shownote     \undefined \def \shownote      #1{#1}          \fi
\ifx \showarticletitle \undefined \def \showarticletitle #1{#1}   \fi
\ifx \showURL      \undefined \def \showURL       {\relax}        \fi
\providecommand\bibfield[2]{#2}
\providecommand\bibinfo[2]{#2}
\providecommand\natexlab[1]{#1}
\providecommand\showeprint[2][]{arXiv:#2}

\bibitem[\protect\citeauthoryear{Barab{\'a}si and Albert}{Barab{\'a}si and
  Albert}{1999}]%
        {barabasi1999emergence}
\bibfield{author}{\bibinfo{person}{Albert-L{\'a}szl{\'o} Barab{\'a}si} {and}
  \bibinfo{person}{R{\'e}ka Albert}.} \bibinfo{year}{1999}\natexlab{}.
\newblock \showarticletitle{Emergence of scaling in random networks}.
\newblock \bibinfo{journal}{{\em science\/}} \bibinfo{volume}{286},
  \bibinfo{number}{5439} (\bibinfo{year}{1999}), \bibinfo{pages}{509--512}.
\newblock


\bibitem[\protect\citeauthoryear{Broquet, Angelone, Jaquiery, Joly, LENA,
  Lengagne, Plenet, Luquet, and Perrin}{Broquet et~al\mbox{.}}{2010}]%
        {broquet2010genetic}
\bibfield{author}{\bibinfo{person}{Thomas Broquet}, \bibinfo{person}{Sonia
  Angelone}, \bibinfo{person}{Julie Jaquiery}, \bibinfo{person}{Pierre Joly},
  \bibinfo{person}{JEAN-PAUL LENA}, \bibinfo{person}{Thierry Lengagne},
  \bibinfo{person}{Sandrine Plenet}, \bibinfo{person}{Emilien Luquet}, {and}
  \bibinfo{person}{Nicolas Perrin}.} \bibinfo{year}{2010}\natexlab{}.
\newblock \showarticletitle{Genetic bottlenecks driven by population
  disconnection}.
\newblock \bibinfo{journal}{{\em Conservation Biology\/}} \bibinfo{volume}{24},
  \bibinfo{number}{6} (\bibinfo{year}{2010}), \bibinfo{pages}{1596--1605}.
\newblock


\bibitem[\protect\citeauthoryear{Erdos, R{\'e}nyi, et~al\mbox{.}}{Erdos
  et~al\mbox{.}}{1960}]%
        {erdos1960evolution}
\bibfield{author}{\bibinfo{person}{Paul Erdos}, \bibinfo{person}{Alfr{\'e}d
  R{\'e}nyi}, {et~al\mbox{.}}} \bibinfo{year}{1960}\natexlab{}.
\newblock \showarticletitle{On the evolution of random graphs}.
\newblock \bibinfo{journal}{{\em Publ. Math. Inst. Hung. Acad. Sci\/}}
  \bibinfo{volume}{5}, \bibinfo{number}{1} (\bibinfo{year}{1960}),
  \bibinfo{pages}{17--60}.
\newblock


\bibitem[\protect\citeauthoryear{Holland}{Holland}{1992}]%
        {holland1992genetic}
\bibfield{author}{\bibinfo{person}{John~H Holland}.}
  \bibinfo{year}{1992}\natexlab{}.
\newblock \showarticletitle{Genetic algorithms}.
\newblock \bibinfo{journal}{{\em Scientific american\/}} \bibinfo{volume}{267},
  \bibinfo{number}{1} (\bibinfo{year}{1992}), \bibinfo{pages}{66--73}.
\newblock


\bibitem[\protect\citeauthoryear{Huizinga, Mouret, and Clune}{Huizinga
  et~al\mbox{.}}{2016}]%
        {huizinga2016does}
\bibfield{author}{\bibinfo{person}{Joost Huizinga},
  \bibinfo{person}{Jean-Baptiste Mouret}, {and} \bibinfo{person}{Jeff Clune}.}
  \bibinfo{year}{2016}\natexlab{}.
\newblock \showarticletitle{Does aligning phenotypic and genotypic modularity
  improve the evolution of neural networks?}. In \bibinfo{booktitle}{{\em
  Proceedings of the Genetic and Evolutionary Computation Conference 2016}}.
  \bibinfo{pages}{125--132}.
\newblock


\bibitem[\protect\citeauthoryear{Lu, Whalen, Boddeti, Dhebar, Deb, Goodman, and
  Banzhaf}{Lu et~al\mbox{.}}{2018}]%
        {lu2018nsga}
\bibfield{author}{\bibinfo{person}{Zhichao Lu}, \bibinfo{person}{Ian Whalen},
  \bibinfo{person}{Vishnu Boddeti}, \bibinfo{person}{Yashesh Dhebar},
  \bibinfo{person}{Kalyanmoy Deb}, \bibinfo{person}{Erik Goodman}, {and}
  \bibinfo{person}{Wolfgang Banzhaf}.} \bibinfo{year}{2018}\natexlab{}.
\newblock \showarticletitle{Nsga-net: a multi-objective genetic algorithm for
  neural architecture search}.
\newblock  (\bibinfo{year}{2018}).
\newblock


\bibitem[\protect\citeauthoryear{Miikkulainen and Forrest}{Miikkulainen and
  Forrest}{2021}]%
        {miikkulainen2021biological}
\bibfield{author}{\bibinfo{person}{Risto Miikkulainen} {and}
  \bibinfo{person}{Stephanie Forrest}.} \bibinfo{year}{2021}\natexlab{}.
\newblock \showarticletitle{A biological perspective on evolutionary
  computation}.
\newblock \bibinfo{journal}{{\em Nature Machine Intelligence\/}}
  \bibinfo{volume}{3}, \bibinfo{number}{1} (\bibinfo{year}{2021}),
  \bibinfo{pages}{9--15}.
\newblock


\bibitem[\protect\citeauthoryear{Owais and Osman}{Owais and Osman}{2018}]%
        {owais2018complete}
\bibfield{author}{\bibinfo{person}{Mahmoud Owais} {and}
  \bibinfo{person}{Mostafa~K Osman}.} \bibinfo{year}{2018}\natexlab{}.
\newblock \showarticletitle{Complete hierarchical multi-objective genetic
  algorithm for transit network design problem}.
\newblock \bibinfo{journal}{{\em Expert Systems with Applications\/}}
  \bibinfo{volume}{114} (\bibinfo{year}{2018}), \bibinfo{pages}{143--154}.
\newblock


\bibitem[\protect\citeauthoryear{Stanley and Miikkulainen}{Stanley and
  Miikkulainen}{2002}]%
        {stanley2002evolving}
\bibfield{author}{\bibinfo{person}{Kenneth~O Stanley} {and}
  \bibinfo{person}{Risto Miikkulainen}.} \bibinfo{year}{2002}\natexlab{}.
\newblock \showarticletitle{Evolving neural networks through augmenting
  topologies}.
\newblock \bibinfo{journal}{{\em Evolutionary computation\/}}
  \bibinfo{volume}{10}, \bibinfo{number}{2} (\bibinfo{year}{2002}),
  \bibinfo{pages}{99--127}.
\newblock


\bibitem[\protect\citeauthoryear{Vasques~Filho and O'Neale}{Vasques~Filho and
  O'Neale}{2020}]%
        {vasques2020transitivity}
\bibfield{author}{\bibinfo{person}{Demival Vasques~Filho} {and}
  \bibinfo{person}{Dion~RJ O'Neale}.} \bibinfo{year}{2020}\natexlab{}.
\newblock \showarticletitle{Transitivity and degree assortativity explained:
  The bipartite structure of social networks}.
\newblock \bibinfo{journal}{{\em Physical Review E\/}} \bibinfo{volume}{101},
  \bibinfo{number}{5} (\bibinfo{year}{2020}), \bibinfo{pages}{052305}.
\newblock


\bibitem[\protect\citeauthoryear{Vi{\'e}}{Vi{\'e}}{2019}]%
        {vie2019information}
\bibfield{author}{\bibinfo{person}{Aymeric Vi{\'e}}.}
  \bibinfo{year}{2019}\natexlab{}.
\newblock \showarticletitle{Information Selection Efficiency in Networks: A
  Neurocognitive-Founded Agent-Based Model}.
\newblock In \bibinfo{booktitle}{{\em Network Theory and Agent-Based Modeling
  in Economics and Finance}}. \bibinfo{publisher}{Springer},
  \bibinfo{pages}{11--34}.
\newblock


\bibitem[\protect\citeauthoryear{Vie}{Vie}{2020}]%
        {vie2020qualities}
\bibfield{author}{\bibinfo{person}{Aymeric Vie}.}
  \bibinfo{year}{2020}\natexlab{}.
\newblock \showarticletitle{Qualities, challenges and future of genetic
  algorithms: a literature review}.
\newblock \bibinfo{journal}{{\em arXiv preprint arXiv:2011.05277\/}}
  (\bibinfo{year}{2020}).
\newblock


\end{thebibliography}

\end{document}